\PassOptionsToPackage{unicode}{hyperref}
\PassOptionsToPackage{hyphens}{url}
\documentclass[
]{article}
\usepackage{amsmath,amssymb}
\usepackage{lmodern}
\usepackage{iftex}
\ifPDFTeX
  \usepackage[T1]{fontenc}
  \usepackage[utf8]{inputenc}
  \usepackage{textcomp} % provide euro and other symbols
\else % if luatex or xetex
  \usepackage{unicode-math}
  \defaultfontfeatures{Scale=MatchLowercase}
  \defaultfontfeatures[\rmfamily]{Ligatures=TeX,Scale=1}
\fi
\IfFileExists{upquote.sty}{\usepackage{upquote}}{}
\IfFileExists{microtype.sty}{% use microtype if available
  \usepackage[]{microtype}
  \UseMicrotypeSet[protrusion]{basicmath} % disable protrusion for tt fonts
}{}
\makeatletter
\@ifundefined{KOMAClassName}{% if non-KOMA class
  \IfFileExists{parskip.sty}{%
    \usepackage{parskip}
  }{% else
    \setlength{\parindent}{0pt}
    \setlength{\parskip}{6pt plus 2pt minus 1pt}}
}{% if KOMA class
  \KOMAoptions{parskip=half}}
\makeatother
\usepackage{xcolor}
\usepackage{graphicx}
\makeatletter
\def\maxwidth{\ifdim\Gin@nat@width>\linewidth\linewidth\else\Gin@nat@width\fi}
\def\maxheight{\ifdim\Gin@nat@height>\textheight\textheight\else\Gin@nat@height\fi}
\makeatother
\setkeys{Gin}{width=\maxwidth,height=\maxheight,keepaspectratio}
\makeatletter
\def\fps@figure{htbp}
\makeatother
\usepackage[normalem]{ulem}
\ifLuaTeX
  \usepackage{selnolig}  % disable illegal ligatures
\fi
\IfFileExists{bookmark.sty}{\usepackage{bookmark}}{\usepackage{hyperref}}
\IfFileExists{xurl.sty}{\usepackage{xurl}}{} % add URL line breaks if available
\hypersetup{
  hidelinks,
  pdfcreator={LaTeX via pandoc}}

\author{}
\date{}

\begin{document}

Applications of Federated Learning in IoT for Hyper Personalisation

Dosi Veer

\emph{Abstract: Billions of IoT devices are being deployed, taking
advantage of faster internet, and the opportunity to access more
endpoints. Vast quantities of data are being generated constantly by
these devices but are not effectively being utilised. Using FL training
machine learning models over these multiple clients without having to
bring it to a central server. We explore how to use such a model to
implement ultra levels of personalization unlike before.}

\textbf{I. INTRODUCTION}

In today's world, smart IoT devices are found to be numerous places all
of which communicate with each other through technologies such as WiFi
and bluetooth constantly collecting and transmitting data to each other
and to the cloud and the amount and type of data is humongous. But this
data is currently not being utilised to its fullest potential which I
would like to explore in this paper. IoT devices have sensors and
actuators that are constantly in process and the information of which is
used to create a local ecosystem or web of devices for the user to make
his life hyper personalised and comfortable in a manner unlike never
before. But right now we don't exactly have a use case for what to do
with so much data apart from showing it to the person and deciding what
to do with it or even ignoring it.

IoT is already on the forefront, so we use federated learning for
continuous training of a model on user side so there is a blueprint on
the cloud which is locally available on the users side and gets update
using data such as amazon alexa, smart watches, mobile phone usage,
smart TVs, smart household machines and so on to give relevant answers
in real time as well as hyper personalised suggestions to the user about
their day to day tracking and life in a way unlike ever before. Data is
constantly crunched in and the model being made better with user
choices, So there is an AI that knows everything about them and is
linked together to all other devices to form a complete local web of
their own that is security protected 256 bits.

\emph{Federated Learning (FL) is a concept first introduced by Google in
2016, in which multiple devices collaboratively learn a machine learning
model without sharing their private data under the supervision of a
central server. Federated Learning has primarily four main steps:}

\emph{• Client Selection/Sampling : Server either randomly picks desired
participants from a pool of devices or uses some algorithm for client
selection.}

\emph{• Parameter Broadcasting: Server broadcasts the global model
parameters to the selected clients}

\emph{• Local Model Training: The clients will parallelly retrain the
models using their local data.}

\emph{• Model Aggregation: Clients will send back their local model
parameters to the server and model parameters will be aggregated towards
the global model.} {[}1{]}

\textbf{II. IoT DEVICES}

Internet of Things. IoT devices, characterised by their ability to
collect, transmit, and receive data, have gained significant prominence
across various domains.

These include smart watches, medical wearables, smart refrigerators,
devices that are connected to each other to form a local web and have
edge computing functionalities.

Data processing occurs on these devices or in nearby edge servers. This
local processing can involve data filtering, aggregation, analysis, and
even running machine learning models. This gives way to two positives:

\begin{enumerate}
\def\labelenumi{\arabic{enumi}.}
\item
  \begin{quote}
  Low Latency: Edge computing is designed to minimise latency, ensuring
  that data processing and decision-making occur quickly, which is
  crucial for applications like autonomous vehicles, robotics, and
  real-time monitoring.
  \end{quote}
\item
  \begin{quote}
  Bandwidth Optimization: By processing data locally, edge computing
  reduces the need for transmitting large volumes of data over the
  network. This helps optimise bandwidth usage and lowers data transfer
  costs.
  \end{quote}
\end{enumerate}

Given below are eidfferent types of IoT devices that share the various
types of data including numbers, images, videos and so on.

\includegraphics[width=5.53936in,height=3.98071in]{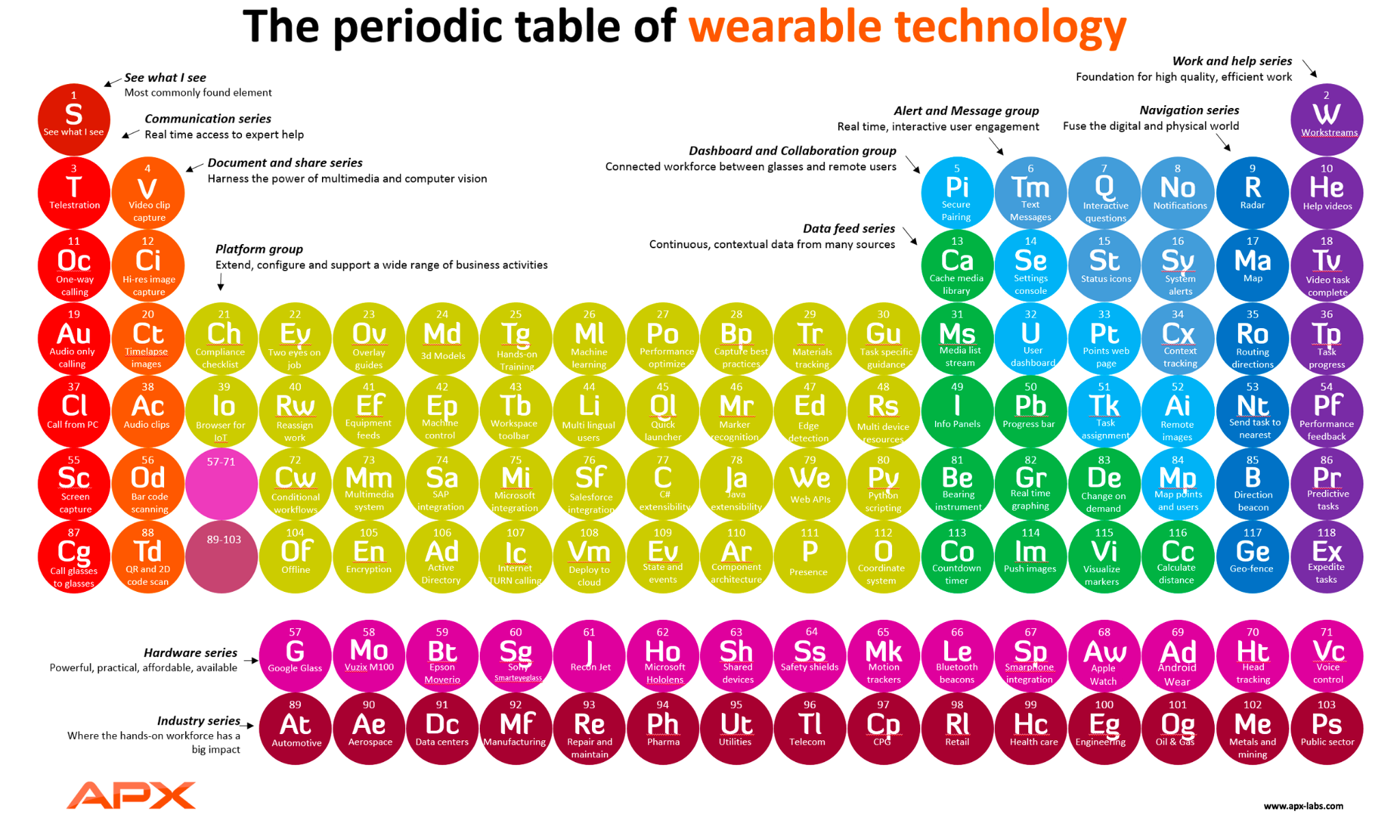}

\begin{quote}
\emph{Figiure 1}

\emph{Source and complete table
\href{http://www.apx-labs.com/landing/periodictable/}{here}}
\end{quote}

\textbf{III. USE CASES:}

A. Smart homes:

For example, a thermostat learns the heating and cooling preferences of
a particular room. Model Updates: Periodically, these locally trained
models send updates (model parameters) to a central server without
sharing raw data. These updates are aggregated to create a global model
that captures insights from all devices.

Smart homes use sensors for security purposes, such as motion detectors,
cameras, and door/window sensors, to monitor and react to events. Model
Updates: These models periodically send updates about detected threats
or anomalies to a central server, without sharing the raw sensor data.
Global Threat Detection: The central server aggregates updates from all
devices, improving the global threat detection model by learning from
the diverse range of data.

\begin{enumerate}
\def\labelenumi{\arabic{enumi}.}
\item
  \begin{quote}
  Wearable technologies (smart watches, smart rings): track health data
  and share that with the central system and have this data collated and
  keep track of anomalies and take the appropriate actions including
  ordering medicines or booking doctors using LLM while keeping medical
  data completely secure. Develop interfaces to enable communication
  between family doctors and the data such as heartbeat and blood
  pressure and sleep being used to generate insights that help doctors
  and caretakers even in some cases.
  \end{quote}
\item
  \begin{quote}
  Smart TV: depending on what you are watching adjust lighting and
  music, order takeaway. Help select better content on a local level.
  \end{quote}
\item
  \begin{quote}
  Home automation systems that respond to voice commands (you can set
  specific words as triggers) and adapt to everyday activities
  \end{quote}
\item
  \begin{quote}
  Apple vision pro: you can create life like scenarios from day to day
  actions
  \end{quote}
\item
  \begin{quote}
  Smart locks, smart refrigerators (that automatically order groceries
  when finished depending on next week's cooking requirements and so
  on), smart air purifiers, smart home robots.
  \end{quote}
\end{enumerate}

We aim to use FL to improve how all such devices interact with each
other in real time as they learn different types of data and activities
from different devices to make it hyper personalised from getting the
morning coffee at the perfect time by syncing up with the alarm to the
right type of music being played at the right time depending on heart
rate levels.

B. Healthcare:

There are a variety of medical devices that are constantly used in
hospitals and nowadays wearable technologies have a lot of ways to track
health but that data needs to be read and understood by a human to give
the correct advice or plan but with ML.

Customised Alerts: Federated learning models can identify patterns and
anomalies in patient data at the individual level. Patients can receive
customised alerts based on their specific health conditions and risk
factors. Personalised Recommendations: The model can offer personalised
recommendations for medication adherence, diet, exercise, and lifestyle
changes, taking into account the patient\textquotesingle s unique health
history and preferences. Treatment Plan Optimization:
Hyper-personalization can optimise treatment plans by considering
real-time patient data. For instance, for a diabetic patient, the system
can adjust insulin dosages based on recent glucose readings. Early
Detection of Health Issues: The model can learn to detect early signs of
health deterioration, enabling timely interventions. For example, it can
predict potential cardiac events in heart patients.

IoT Sensors: IoT sensors can be embedded in medical devices or wearables
to monitor a patient\textquotesingle s condition and treatment progress
continuously. Medication Adherence: Smart pill dispensers can ensure
patients take their medications as prescribed, and IoT-connected
inhalers can track usage for respiratory conditions.

Dose Adjustment: For medication-based treatments, the system can provide
dose adjustments based on real-time health data, minimising side effects
and optimising effectiveness. Behavioural Support: IoT devices can
provide personalised behavioural support, such as reminders,
motivational messages, and exercise routines tailored to the
patient\textquotesingle s preferences and capabilities. Patient
Engagement: Personalised treatment plans are more engaging, increasing
patient compliance and overall treatment success.

C. Personalized Commute Recommendations:

Federated learning can provide personalised commute recommendations to
individuals, suggesting the best modes of transportation and routes
based on their preferences and real-time conditions collected from data
from the IoT sensors available such as those on the road and inside
cameras measuring congestion, vehicle flow and accidents including but
not restricted to,

\begin{enumerate}
\def\labelenumi{\arabic{enumi}.}
\item
  \begin{quote}
  Individualised Routing: Citizens receive personalised route
  recommendations, taking into account their preferred mode of
  transportation, real-time traffic, and historical travel patterns.
  \end{quote}
\item
  \begin{quote}
  Predictive Alerts: Hyper-personalization can send commuters alerts
  about potential delays, route changes, or alternative transportation
  options tailored to their specific needs.
  \end{quote}
\end{enumerate}

Similarly FL can solve energy management by automatically adjusting for
sustainable energy usga eo light and sound as required depending on
inputs being processed through various devices

\textbf{IV. FEDERATED LEARNING}

Federated Learning is all about constantly adapting a model and making
it personalised based on individuals data while keeping data secure.

Ideal problems for federated learning have the following properties: 1)
Training on real-world data from mobile devices provides a distinct
advantage over training on proxy data that is generally available in the
data centre. 2) This data is privacy sensitive or large in size
(compared to the size of the model), so it is preferable not to log it
to the data centre purely for the purpose of model training (in service
of the focused collection principle). 3) For supervised tasks, labels on
the data can be inferred naturally from user interaction. {[}2{]}

The type of federated learning that will work best in the case of IoT
web would be cross device federated. In cross-device FL, clients are
small distributed entities (e.g., smartphones, wearables, and edge
devices), and each client is likely to have a relatively small amount of
local data. Hence, for cross-device FL to succeed, it usually requires a
large number (e.g., up to millions) of edge devices to participate in
the training process. {[}3{]} And right now a lot of data of such
devices is available to big tech corporations who can use it to train an
LLM on a general solution which then constantly learns and improves to
make it seriously customised as more and more such local data about the
user's choices, habits and activities is collected and amortised to make
a viable input and thus better outputs from the LLM.

Cross-device federated learning (FL) intends to use a single distributed
customer group for high-quality machine learning model training while
retaining customer data locally for privacy protection.

\includegraphics[width=4.74323in,height=3.45833in]{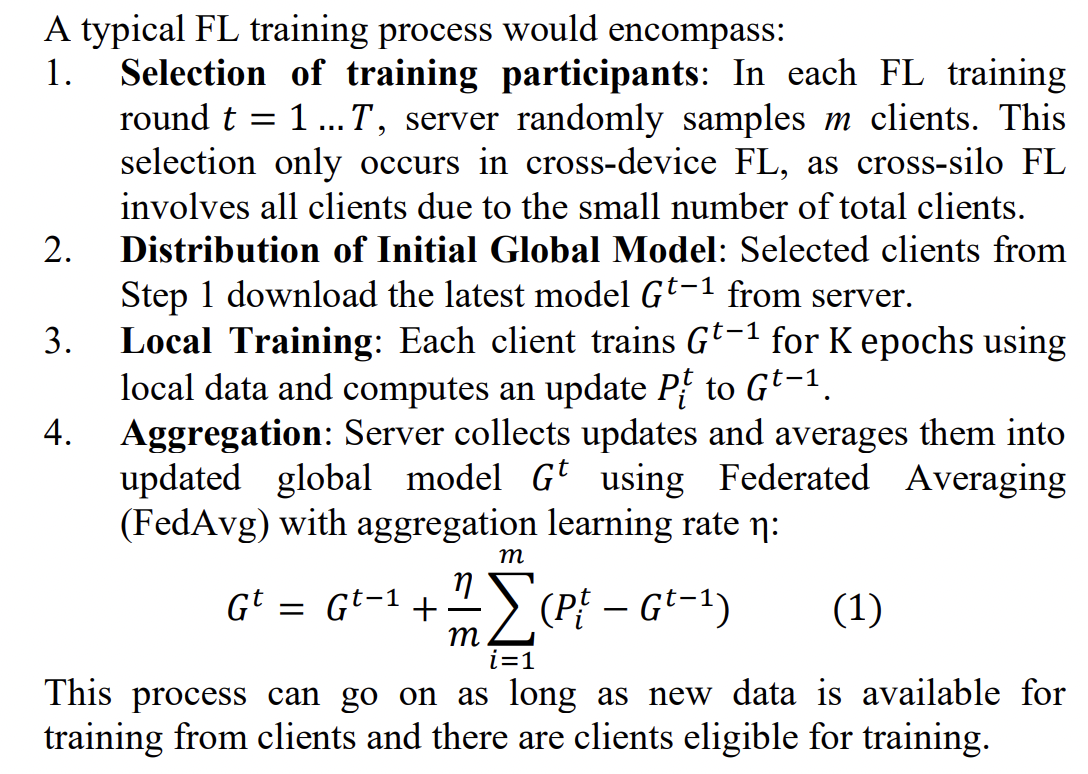}

Figure 2 {[}4{]}

\textbf{V. HETEROGENEITY OF DATA}

Existing federated learning problems are because when executed in
controlled environments data is homogeneous in nature but for real world
applications, the IoT devices that we are typing to link up and train
the model on are highly variable and dynamically changing. Scaling
heterogeneous devices requires researchers to deal with diverse device
hardware and software environments to reflect real computing capacities,
and build distributed communications through the network interface cards
to reflect real transmission capabilities {[}5{]}. So a good way to
solve this is by creating a \emph{FSReal containing a flexible
heterogeneous device runtime, a group of efficient and scalable FL
device executors and FL server, and implementations for a number of
practical and advanced FL enhancement techniques with easy extensions.}
{[}6{]}

The solution to this is An efficient and scalable system which has
diversified device data runtime and modern practicality procedures that
include but are not restricted to customization, data compression. A
cross-device FL system should have good usability and can conduct
evaluation for a wide range of device runtimes. Besides, devices often
have limited resources, such as computing capacity, communication
bandwidth, and storage. {[}7{]}

\textbf{VI. SECURITY OF FEDERATED LEARNING}

Federated learning, which allows a decoupling of data provision at
enduser equipment (UE) and machine learning model aggregation, such as
network parameters of deep learning, at a centralized server. {[}8{]}.
FL in itself is one of the best privacy preserving type of learning
possible since it allows machine learning at source itself instead of
requiring a centre for the data to be transferred over to.

But this is not to say that tier are no vulnerabilities associated with
the same. There can be on model attacks including data poisoning
(manipulation of client data), model posioning, backdoor attacks,
andevasion atacks (manipulating user data to circumvent the program).
Also inference attacks and client dropping out are all security issues
that need to be considered and resolved lest the model become unusable,
biased or worse hacked to reveal data in raw form.

There may also be a compromised server, gradient leakage or just plain
non malicious failure which can contribute to being as kink in the
armour of FL.

\textbf{VII. EFFICIENCY OF FEDERATED LEARNING}

Federated learning excels in two major characteristics that are of high
importance to use in the given context which is data privacy and real
time training and implementation of the model. Federated Learning excels
in preserving data privacy, as it allows model training without
transferring raw data. And FL can train the ML model regularly as new
data is fed to it thus making it better and more capable. However, the
privacy-preserving techniques employed, such as differential privacy or
encryption, and the rapid and constant training that is required
introduces computational overhead as compared to supervised ML training
and thus making it require more energy and calculative power.

\textbf{Efficiency metrics:}

\begin{enumerate}
\def\labelenumi{\arabic{enumi}.}
\item
  \begin{quote}
  Communication Overhead: One of the primary benefits of Federated
  Learning is its reduced communication overhead compared to traditional
  centralised approaches, which can be a game-changer in low-bandwidth
  or high-latency settings.
  \end{quote}
\item
  \begin{quote}
  Computational Efficiency: Federated Learning can distribute
  computation across devices, potentially saving computational resources
  and energy.
  \end{quote}
\item
  \begin{quote}
  Scalability: The efficiency of Federated Learning systems in scaling
  to large numbers of devices or participants is a critical
  consideration.
  \end{quote}
\end{enumerate}

All of these metrics can be used to find out how effective different
ways of making FL efficient are and how we can analyse and make them
better.

There are a few ways to solve the issue of FL not being ``green''
enough. The first of which is as follows:

A. Configuration of Processing Resource

In Federated Learning, each participating device, also known as a
client, holds a portion of the model\textquotesingle s parameters, which
include kernel code (neural network layers) and weights. These
parameters collectively define the model\textquotesingle s architecture
and its ability to make predictions or classifications. Amortisation:
The term "amortisation" in this context refers to spreading the
computational load associated with model updates and training across
multiple samples or data points. Instead of processing each data point
individually.

Amortisation has the following benefits: improved training efficiency
because of reduced data and instruction fetches. But there are some
considerations to be taken during implementation including:

\begin{enumerate}
\def\labelenumi{\arabic{enumi}.}
\item
  \begin{quote}
  Batch size: The choice of batch size plays a critical role in
  amortisation. A larger batch size amortises resources more effectively
  but may require more memory and computational power on the client
  device.
  \end{quote}
\item
  \begin{quote}
  Communication rounds: model updates are typically aggregated over
  multiple communication rounds. Amortising resources over a batch of
  samples within each round can help distribute the computational load
  more evenly.
  \end{quote}
\item
  \begin{quote}
  Resource constraints: It\textquotesingle s essential to consider the
  computational and memory constraints of client devices.
  \end{quote}
\end{enumerate}

It is quintessential that we keep all of this in mind while looking at
practical implementations since IoT devices have different communication
rounds and batch sizes that need to be adjusted in a way to satisfy all
demands while looking at the resource constraints of each device.

B. Focusing only on change.

For the first frame, the entire frame must be processed. For successive
frames, a ``diff'' can be taken between the current and prior frames,
directing the work only to those portions that changed. Those changes
are categorised as ``events,'' and this kind of architecture is an
``event-based'' approach. Events can represent various data streams such
as change in data or any difference from the ordinary or the daily. By
processing only the changes, an event-based architecture significantly
reduces the computational load compared to processing entire frames.
This is especially advantageous when dealing with real-time or
resource-constrained environments. This can also help a lot with low
latency.

\textbf{VII: ALGORITHMS AVAILABLE}

The algorithm that we would like to consider is Personalised federated
averaging which uses occassional communication with an aggregating
parameter server. All user devices share a set of base layers with same
weights (colored blue) and have distinct personalization layers that can
potentially adapt to individual data. The base layers are shared with
the parameter server while the personalization layers are kept private
by each device {[}9{]}. The quantity of training samples for each user
is insufficient to independently train machine learning models. As a
result, wisdom of the crowd is required. The idea proposed is training
the base layer by federated learning and the second personalisation
layer to be trained with deep downward slope to form a 2 layer deep
neural network

Benefits of Personalized Federated Averaging:

\begin{itemize}
\item
  \begin{quote}
  Privacy-Preserving: Only model updates are shared with the server.
  \end{quote}
\item
  \begin{quote}
  Customization: Clients can tailor the global model to their specific
  users or tasks, resulting in more relevant recommendations or
  services.
  \end{quote}
\item
  \begin{quote}
  Adaptability: The global model evolves over time, improving its
  ability to provide personalised experiences as more clients
  contribute.
  \end{quote}
\end{itemize}

On each client device, the global model is personalised to the
user\textquotesingle s preferences or requirements. This can be achieved
in several ways:

\begin{itemize}
\item
  \begin{quote}
  Fine-Tuning: Clients fine-tune the global model using their local
  data, which may include user-specific preferences, historical
  interactions, or context.
  \end{quote}
\item
  \begin{quote}
  Transfer Learning: Clients use transfer learning techniques to adapt
  the global model\textquotesingle s knowledge to their specific tasks
  or users.
  \end{quote}
\item
  \begin{quote}
  Custom Layers: Clients can add custom layers to the global model to
  capture user-specific features or requirements.
  \end{quote}
\end{itemize}

Clients participate in federated updates by sending their personalised
models (model weights and architecture modifications) to the central
server.

\textbf{VIII. REAL LIFE EXAMPLES}

\textbf{In smart homes,}

Each IoT device in the smart home collects data about user preferences,
habits, and usage patterns without sending this data to a central
server. Instead, federated learning is employed, where the device-level
machine learning models are trained locally on each IoT device. User
Profiles: The smart home system creates user profiles for each household
member based on their interactions with the IoT devices. For example, it
recognizes when each person wakes up, goes to bed, adjusts the
thermostat, or uses specific appliances. Local Model Training: The data
collected by IoT devices is used to train local machine learning models
on the respective devices. Each device\textquotesingle s model learns to
anticipate the preferences and behaviour of the household member
interacting with it. Collaborative Learning: Periodically, the local
models on each IoT device share updates with a central aggregation
server, which aggregates and merges these updates into a global model
without ever directly accessing the raw data. The global model then
captures the collective knowledge of all the devices without
compromising user privacy.

Consider a family of four residing in this smart home:

\begin{itemize}
\item
  \begin{quote}
  Sophia is a fitness enthusiast who wakes up early for her morning jog.
  Her AR glasses monitor her vitals, and her smart kitchen appliances
  prepare a customised post-workout smoothie.
  \end{quote}
\item
  \begin{quote}
  Ethan is a bookworm who enjoys reading in the evening. His AR glasses
  suggest books from his favourite authors, and the lighting and music
  in his reading nook adjust to his preferences.
  \end{quote}
\item
  \begin{quote}
  Ava is a health-conscious vegan with specific dietary requirements.
  Her smart refrigerator alerts her when ingredients are about to expire
  and suggests plant-based recipes.
  \end{quote}
\item
  \begin{quote}
  Noah is a tech-savvy gamer who loves VR experiences. His personal
  robot doubles as a gaming companion and adjusts room lighting and
  temperature for immersive gameplay.
  \end{quote}
\end{itemize}

\textbf{In healthcare,}

Remote Monitoring of Chronic Disease Patients:

Patients use IoT devices such as wearable fitness trackers, smart
glucose metres, or blood pressure monitors to collect health data
regularly. These devices are equipped with machine learning models
capable of processing the data locally.

Instead of sending raw data to a central server, each IoT device adds to
its machine learning model using the collected data. For instance, a
smart insulin pump can update its insulin dosage recommendation model
using the patient\textquotesingle s blood sugar readings. These updates
are aggregated without disclosing the individual patient data, ensuring
privacy.

\textbf{IX. CHALLENGE AND RESOLVING THEM}

\textbf{A. Trustworthiness}

In practical deployment scenarios, IoT devices have become attractive
targets for potential adversaries who aim to carry out various attacks,
such as phishing, identity theft, and distributed denial of service
(DDoS). Despite the fact that these attacks could be mitigated by simply
installing security patches, many IoT devices lack the necessary
computational resources to do so. Furthermore, in a multi-device system,
it is challenging to determine whether a new participant is malicious or
not before it becomes part of the system.\emph{Therefore, a critical
consideration in FL design is to ensure that the server as a black box
for aggregation that does not learn the locally trained model of each
user during model aggregation. to implement lightweight and secure
aggregation protocols that could provide the same level of privacy and
dropout resiliency guarantees while substantially reducing the
aggregation complexity, which meets the constraint in the IoT setting.}
{[}10{]}

\textbf{B. Restricted available network bandwidth:}

Communication restriction is taken as one of the major challenges while
implementing an FL based IoT environment. Today, most of the IoT devices
that are popular and widely used communicate on a network bandwidth of
wireless networks which is much less than wired network bandwidth.

This makes communication problem of two kinds: inefficiency of server
communication and straggler clients which are not able to send back data
in the communication rounds

There are various solutions proposed but we have outlined the top three
of them below:

1. Client-Edge-Cloud Hierarchical Federated Learning {[}11{]}: It allows
for model aggregation at two levels - one at the edge and one at the
cloud. The algorithm converges under certain conditions, and convergence
analysis provides qualitative guidelines on picking the aggregation
frequencies at two levels. The benefits of HiAVG include reduced model
training time and energy consumption of the end devices compared to
traditional cloud-based Federated Learning. Additionally, it allows for
better communication-computation trade-offs compared to cloud-based
systems.

2. Joint Learning and Communications Framework {[}12{]}: The proposed
framework enables the implementation of federated learning algorithms
over wireless networks by jointly considering user selection and
resource allocation to minimise FL training loss. The framework
formulates an optimization problem that takes into account FL and
wireless metrics and factors. The expected convergence rate of the FL
algorithm is derived to determine the optimal transmit power given the
user selection and uplink resource block allocation. The Hungarian
algorithm is used to find the optimal user selection and resource
allocation to minimise the FL loss function. Simulation results show
that the joint federated learning and communication framework yields
significant improvements in performance compared to existing
implementations of the FL algorithm that do not account for the
properties of the wireless channel.

3. Mercury {[}13{]}: three novel techniques: group-wise importance
computation and sampling, importance-aware data resharding, and
bandwidth-adaptive computation-communication scheduling, which fully
exploit the benefits brought by importance sampling.

\textbf{X. CONCLUSION}

Federated Learning based IoT environments will be really useful in this
age where everything from content consumed to posts you see on your feed
being customised so as to take life activities to the next level by
making them hyper personalised in their nature. In this article we
highlighted how we can go about utilising this opportunity to create the
next big wave of personalisation while maintaining its efficiency and
security needs. We have also given the challenges and how they might be
resolved to create a solution while maintaining efficiency and keeping
the solution ``green\textquotesingle\textquotesingle{} enough. We hope
that this article acts as inspiration for future research into the
unique combination of Internet of Things and FL.

References:

{[}1{]} Mammen, P. M. (2021, January 14). \emph{Federated Learning:
opportunities and challenges}. arXiv.org.
\href{https://arxiv.org/abs/2101.05428}{\uline{https://arxiv.org/abs/2101.05428}}

{[}2{]} McMahan, H. B. (2016, February 17).
\emph{Communication-Efficient Learning of Deep Networks from
Decentralized Data}. arXiv.org.
\href{https://arxiv.org/abs/1602.05629}{\uline{https://arxiv.org/abs/1602.05629}}

{[}3{]} Huang, C. (2022, June 26). \emph{Cross-Silo Federated Learning:
Challenges and opportunities}. arXiv.org.
\href{https://arxiv.org/abs/2206.12949}{\uline{https://arxiv.org/abs/2206.12949}}

{[}4{]} Pye, S. K. (2021, August 22). \emph{Personalised Federated
Learning: a combinational approach}. arXiv.org.
\href{https://arxiv.org/abs/2108.09618}{\uline{https://arxiv.org/abs/2108.09618}}

{[}5{]} Qinbin Li, Zeyi Wen, Zhaomin Wu, Sixu Hu, Naibo Wang, Yuan Li,
Xu Liu, and Bingsheng He. 2021. A Survey on Federated Learning Systems:
Vision, Hype and Reality for Data Privacy and Protection. IEEE
Transactions on Knowledge and Data Engineering (2021),

{[}6{]} TensorFlow: Large-Scale Machine Learning on Heterogeneous
Systems. arXiv preprint arXiv:1603.04467 (2015).

{[}7{]} Chen, D. (2023, March 23). \emph{FS-Real: Towards Real-World
Cross-Device Federated learning}. arXiv.org.
\href{https://arxiv.org/abs/2303.13363}{\uline{https://arxiv.org/abs/2303.13363}}

{[}8{]} \emph{On safeguarding privacy and security in the framework of
federated learning}. (2020, August 1). IEEE Journals \& Magazine
\textbar{} IEEE Xplore.
\href{https://ieeexplore.ieee.org/document/9048613}{\uline{https://ieeexplore.ieee.org/document/9048613}}

{[}9{]} Arivazhagan, M. G. (2019, December 2). \emph{Federated Learning
with Personalization Layers}. arXiv.org.
\href{https://arxiv.org/abs/1912.00818}{\uline{https://arxiv.org/abs/1912.00818}}

{[}10{]} Zhang, T. (2021, November 15). \emph{Federated Learning for
Internet of Things: Applications, challenges, and opportunities}.
arXiv.org.
\href{https://arxiv.org/abs/2111.07494}{\uline{https://arxiv.org/abs/2111.07494}}

{[}11{]} L. Liu, J. Zhang, S. Song, and K. B. Letaief,
``Client-edge-cloud hierarchical federated learning,'' 2020

{[}12{]} M. Chen, Z. Yang, W. Saad, C. Yin, H. V. Poor, and S. Cui, ``A
joint learning and communications framework for federated learning over
wireless networks,''

{[}13{]} X. Zeng, M. Yan, and M. Zhang, ``Mercury: Efficient on-device
distributed dnn training via stochastic importance sampling,''

\end{document}